\documentclass[10pt,twocolumn,letterpaper]{article}

\usepackage[pagenumbers]{cvpr} %

\usepackage{float}
\usepackage{multirow}
\usepackage[table]{xcolor}
\usepackage{graphicx}
\usepackage[most]{tcolorbox}

\definecolor{cvprblue}{rgb}{0.21,0.49,0.74}
\usepackage[pagebackref,breaklinks,colorlinks,allcolors=cvprblue]{hyperref}
\usepackage{setspace}
\usepackage{amsmath}
\usepackage{amssymb}
\usepackage{pifont}
\newcommand{\cmark}{\textcolor{green!60!black}{\ding{51}}}
\newcommand{\xmark}{\textcolor{red!70!black}{\ding{55}}}

\def\paperID{}
\def\confName{CVPR}
\def\confYear{2026}

\newcommand{\customfootnotetext}[2]{{%
  \renewcommand{\thefootnote}{#1}%
  \footnotetext[0]{#2}}}%

\title{3D-Layout-R1: Structured Reasoning for Language-Instructed Spatial Editing}

\author{
Haoyu Zhen$^{1,2}$\footnotemark[1]\ \,\footnotemark[2]
\hspace{-2mm}
\and Xiaolong Li$^{1}$\footnotemark[1]
\and Yilin Zhao$^{1}$\footnotemark[1]
\and Han Zhang$^{1}$
\and Sifei Liu$^{1}$
\hspace{-2mm}
\and Kaichun Mo$^{1}$
\vspace{0.01cm}
\and Chuang Gan$^{2}$
\and Subhashree Radhakrishnan$^{1}$
\and
\\
$^1$NVIDIA\hspace{10em}$^2$UMass Amherst
}

\begin{document}

\twocolumn[{
    \renewcommand\twocolumn[1][]{#1}
    \maketitle
    \centering
    \vspace{-2mm}
    \begin{minipage}{0.96\textwidth}
        \centering
        \includegraphics[trim=000mm 000mm 000mm 000mm, clip=False, width=\linewidth]{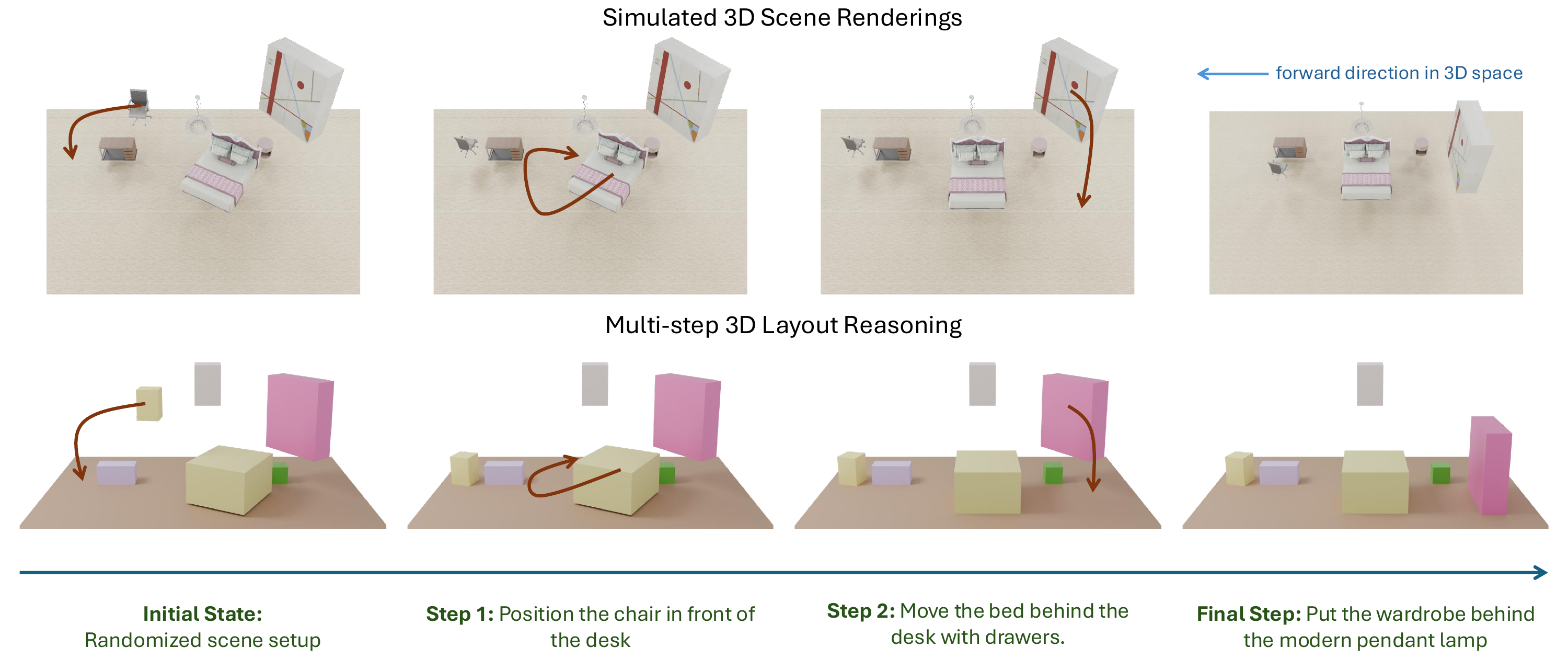}
    \end{minipage}
    \captionsetup{type=figure}
    \captionof{figure}{We introduce \textbf{3D-Layout-R1}, which performs multi-step language-guided 3D layout editing, iteratively updating an initially randomized scene into a sequence of spatially consistent intermediate layouts.} 
    \label{fig:teaser}
    \vspace{2em}
}]

\customfootnotetext{*}{Equal contribution.}
\customfootnotetext{$\dag$}{This work was done while Haoyu Zhen was an intern at NVIDIA.}

\maketitle

\begin{abstract}
Large Language Models (LLMs) and Vision Language Models (VLMs) have shown impressive reasoning abilities, yet they struggle with spatial understanding and layout consistency when performing fine-grained visual editing. We introduce a Structured Reasoning framework that performs text-conditioned spatial layout editing via scene-graph reasoning. Given an input scene graph and a natural-language instruction, the model reasons over the graph to generate an updated scene graph that satisfies the text condition while maintaining spatial coherence. By explicitly guiding the reasoning process through structured relational representations, our approach improves both interpretability and control over spatial relationships. We evaluate our method on a new text-guided layout editing benchmark encompassing sorting, spatial alignment, and room-editing tasks. Our training paradigm yields an average 15\% improvement in IoU and 25\% reduction in center-distance error compared to Chain of thought Fine-tuning (CoT-SFT) and vanilla GRPO baselines. Compared to SOTA zero-shot LLMs, our best models achieve up to 20\% higher mIoU, demonstrating markedly improved spatial precision.

\end{abstract}

\section{Introduction}
Understanding and manipulating 3D scenes through natural language is a fundamental capability for agents and content creation systems. Beyond passive perception, agents must be able to rearrange their surroundings, e.g., “move the chair from the desk to align with the sofa”, which requires a sophisticated blend of capabilities: comprehending compositional spatial relationships, understanding semantic intent, and adhering to strict physical constraints to produce a plausible layout. While recent progress in 3D perception and multimodal foundation models has advanced the ability of VLMs to answer spatial questions, far fewer methods can execute structured and multi-step 3D layout edits in response to natural language instructions.

Existing VLM-based spatial reasoning frameworks have primarily focused on passive 3D understanding. Models like SpatialRGPT~\citep{cheng2024spatialrgpt}, SpatialLLM~\citep{ma2025spatialllm}, SpatialReasoner~\citep{spatialreasoner}, and 3D-R1~\citep{huang20253d} enhance spatial VQA through depth cues, implicit 3D representations, or coordinate-based reasoning. Despite these gains, such systems do not modify the underlying 3D scene and lack the structured mechanisms needed for long-horizon editing. This gap motivates a shift from answering spatial queries to acting upon 3D layouts in an interpretable and physically consistent manner.

Language-driven 3D layout editing has recently emerged as a promising direction. A recent line of works use LLMs as high-level planners to edit existing layouts. In this dominant paradigm, the VLM generates a high-level plan or a set of spatial constraints, which is then passed to a separate, external module, such as a constraint solver~\citep{yang2024holodeck} or a differentiable optimizer~\citep{sun2025layoutvlm, el2025scanedit}, to compute the final 3D poses. While effective for ensuring physical feasibility, these pipelines are limited by manually specified rules or objectives, reduced flexibility for diverse instructions, difficulty handling long-horizon compositional edits, and a largely opaque reasoning process that offers little interpretability or opportunity for correction. Other methods perform one-shot layout prediction or single-object placement~\cite{abdelreheem2025placeit3d}, but these do not generalize to multi-object rearrangement or sequential editing of existing scenes. To the best of our knowledge, no existing system supports fully integrated, multi-step 3D layout editing that directly reasons over structured spatial representations.

In this work, we introduce \textbf{3D-Layout-R1}, a framework that performs structured, interpretable, and language-directed 3D layout editing by reasoning directly over a 3D bounding-box based scene graph as an iterative canvas. Instead of generating a vague, free-form chain-of-thought (CoT), our model, 3D-Layout-R1, produces a structured trace of scene-graph transformations (Fig~\ref{fig:teaser}). Each reasoning step is an explicit, verifiable graph edit that directly updates the scene's state. This approach embeds the 3D spatial logic directly within the model's generation process. This allows 3D-Layout-R1 to plan and execute complex, multi-step rearrangements (e.g., "first move the box, then place the lamp next to the book") while ensuring each intermediate step is interpretable and geometrically coherent.

\begin{figure*}[tbp]
    \centering
    \includegraphics[width=0.98\linewidth]{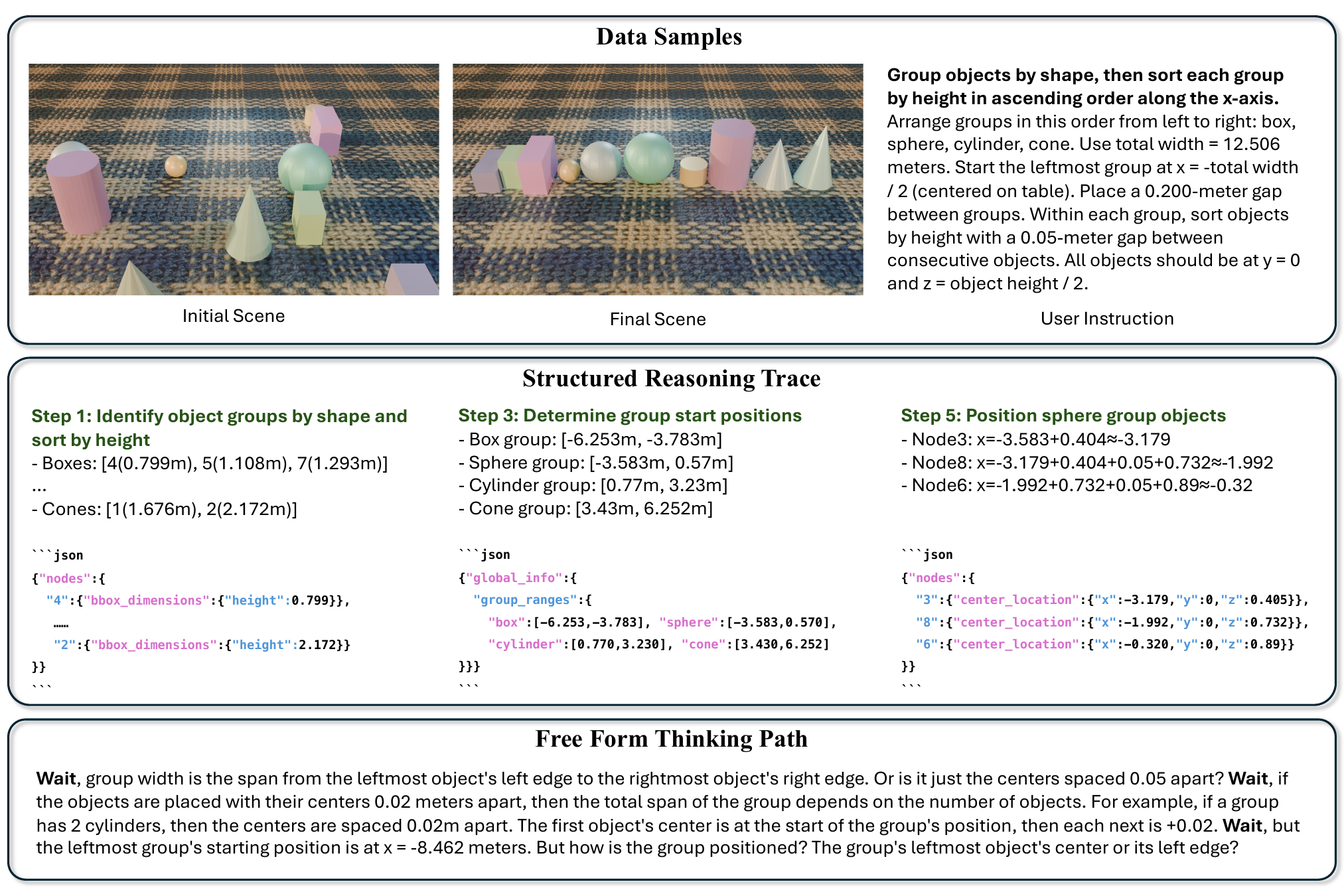}
    \caption{\textbf{Example from the synthetic 3D sorting benchmark.} Given a instruction to group and sort objects by shape and height, our model generates a concise structured reasoning trace with JSON scene-graph updates that transforms the initial scene into the final layout, in contrast to a long, ambiguous free-form thinking path.}
    \vspace{-3mm}
    \label{fig:sorting_qual}
\end{figure*}

To achieve this, we integrate a GRPO-based reinforcement learning stage that optimizes layout accuracy using a dense 3D IoU reward and collision-aware penalties. By jointly leveraging structured scene-graph reasoning and RL-driven refinement, the model learns to generate precise, physically consistent layout edits that reliably satisfy complex textual instructions.

\noindent Our contributions are summarized as follows:
\begin{enumerate}
    \item 
    We introduce \textbf{3D-Layout-R1}, a framework that directly performs multi-step 3D scene editing over a 3D-Layout using language-guided, interpretable chain-of-graph-edits reasoning across three tasks: Sorting, Spatial Alignment, and Room Editing.
    \item 
    We release a dataset of 15k 3D scenes with natural language instructions, intermediate chain-of-thought graph edits, and target layouts, providing the first benchmark dedicated to multi-step editing of existing 3D scenes.
    \item 
    We design geometric reward functions based on 3D IoU for policy optimization and evaluation, offering continuous and geometrically faithful supervision than distance-based or binary success metrics.
    \item 
    3D-Layout-R1 improves mean IoU by \textbf{15\%} and reduces center-distance errors by \textbf{25--30\%} compared to CoT-SFT and GRPO baselines. It achieves up to \textbf{20\%} higher IoU compared to leading zero-shot LLMs and prior spatial reasoning or layout editing systems.
\end{enumerate}

\section{Related Works}
\textbf{Spatial Reasoning.}
Recent efforts to endow VLMs with spatial intelligence have explored diverse modalities and representations. Several approaches enhance 2D VLMs with 3D positional cues from RGB-D inputs or multi-view images~\citep{cheng2024spatialrgpt, cai2025spatialbot, zhu2025llava, cheng2025sr3d, daxberger2025mm}, while others learn implicit “cognitive maps” from video~\citep{zheng2025video, yang2025thinking} or ground reasoning in 2D coordinates and large-scale embodied datasets~\citep{liu2025spatialcot, song2025robospatial, zhou2025roborefer, sarch2025grounded}.
The most proximate works leverage explicit 3D representations for spatial reasoning. SpatialVLM~\citep{chen2024spatialvlm} taught metric relationships by extracting 3D properties from 2D images, while SpatialRGPT~\citep{cheng2024spatialrgpt} generated region-aware Visual Question Answering (VQA) using implicit 3D scene graphs. Furthermore,~\citep{ma2025spatialllm} and ~\citep{ma2025spatialreasoner} incorporated estimated depth, distances, and explicit 3D coordinates to enable multi-step reasoning. In contrast to these works, which primarily focus on spatial VQA using implicit or coordinate-based representations, we investigate whether LLMs/VLMs can reason and update over the structure of a 3D scene graphs.  

\noindent
\textbf{Language-Driven 3D Layout Editing.}
Traditional 3D scene editing requires expert knowledge and manual operation. Recent language-driven approaches aim to automate spatial manipulation using LLMs/VLMs. Some methods operate via intermediate representations, such as \cite{zhou2024layout, lin2024instructscene}, which either edit in 2D before lifting to 3D or construct semantic graph priors for scene decoding. Others model 3D structure directly, including diffusion- or placement-based techniques \cite{zheng2024editroom, abdelreheem2025placeit3d}, though these are often restricted to simple object-level edits. A complementary line of work uses LLMs as \emph{high-level planners} that produce spatial constraints or initial poses for external optimizers \cite{feng2023layoutgpt, yang2024holodeck, el2025scanedit, sun2025layoutvlm}, or leverage symbolic and agentic pipelines for procedural scene manipulation \cite{gu2025blendergym, huang2025fireplace}. More recent efforts explore direct 3D object generation with supervised or DPO-tuned LLMs \cite{yang2025optiscene}. While these approaches demonstrate the promise of natural-language 3D editing, most rely on external solvers and struggle with complex compositional instructions. Concurrent work such as \cite{ran2025direct} explores end-to-end layout reasoning in 2D, whereas we focus on enabling a LLM to directly reason over and iteratively update a structured 3D scene graph for long-horizon compositional editing without external optimization.

\noindent
\textbf{Structured Chain-of-Thought.}
Recent advances in reasoning~\cite{Guo2025deepseekr1, qwen3} highlight that while Chain-of-thought (CoT) reasoning is powerful, unconstrained generation from standard RL methods~\cite{deepseek-math, yu2025dapo, gspo} can produce incoherent or hallucinated reasoning traces~\cite{huang2025survey}. To mitigate this, researchers have introduced structured constraints into the reasoning process, enhancing performance on tasks requiring complex, structured outputs~\cite{ling2023deductive, li2025RaLU, Li25Structured}.
Extending this to the multimodal domain, recent works improve model performance by leveraging explicit visual meta-information~\cite{huang2025frag,Anurag2025Temporal,guo2025structuredoutputsenablegeneralpurpose} or adopting specialized modules for complex spatio-temporal graphs~\cite{Fei2024Video}. However, directly integrating structured 3D information into the reasoning process of LLMs/VLMs is less explored. We posit that embedding our 3D bounding-box scene graph within the reasoning trace guides more faithful spatial logic, mitigates multimodal hallucination, and promotes more accurate, generalizable reasoning.

\section{Method}

\subsection{Preliminaries}
\paragraph{Scene graph representation.}
\label{para:scene-graph-representation}
We represent each 3D layout as a directed scene graph with nodes corresponding to objects and
supporting regions, and edges encoding contact or containment relations. In practice, each graph is
serialized as a JSON dictionary keyed by integer node identifiers, where each node stores
attributes such as \texttt{node\_type},
\texttt{center\_location} (a 3D position), \texttt{dimension} (axis-aligned length, width, height),
\texttt{rotation} (roll, pitch, yaw), and an optional natural-language \texttt{caption}.

\begin{figure*}[t]
    \centering
    \includegraphics[width=0.98\linewidth]{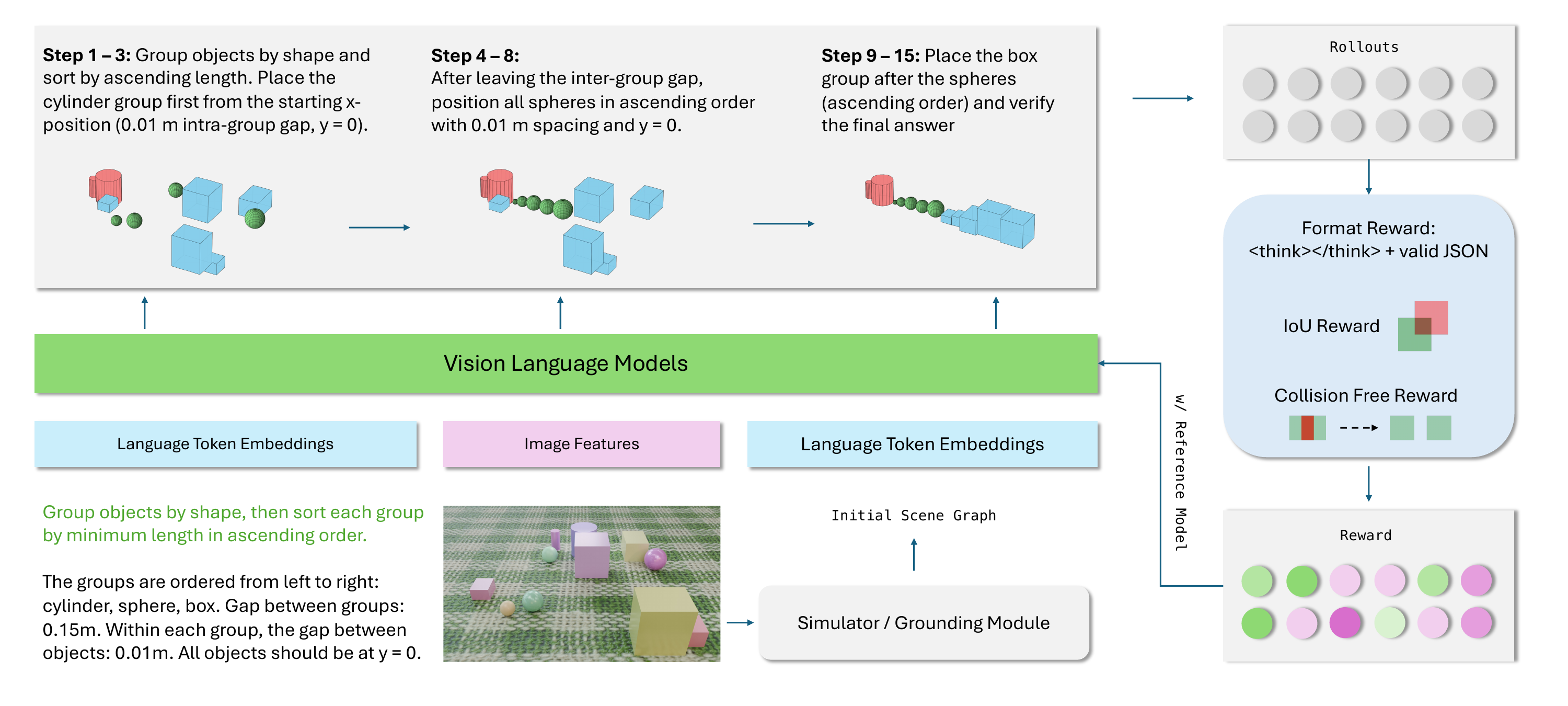}
    \vspace{-5mm}
    \caption{\textbf{Overview of our training pipeline.} The vision-language model predicts step-by-step layout edits from the instruction and initial scene graph, and rollouts are optimized using a combination of format, IoU, and collision-free rewards.}
    \vspace{-3mm}
    \label{fig:placeholder}
\end{figure*}

\paragraph{CoT-SFT and GRPO Training.} Modern approaches to enhancing reasoning capabilities in large language models~\cite{Guo2025deepseekr1} typically follow a two-stage pipeline: CoT-SFT followed by RL optimization. During the RL stage, recent methods~\cite{deepseek-math} leverage a policy optimization framework that enables the model to sample multiple candidate outputs and update its parameters via a clipped surrogate objective:
\begin{equation}
\footnotesize
\begin{split}
&\mathcal{J}_{\mathrm{GRPO}}(\theta)
= \mathbb{E}_{q \sim P(Q),\, \{o_i\}_{i=1}^G \sim \pi_{\theta_{\mathrm{old}}}(O \mid q)}
\Bigg[
    \frac{1}{G} \sum_{i=1}^G \frac{1}{|o_i|}
    \\&
    \sum_{t=1}^{|o_i|}
    \Bigg\{
        \min\Big(
            r_{i,t} \hat{A}_{i,t},
            \text{clip}\big( r_{i,t}, 1 - \epsilon, 1 + \epsilon \big)\hat{A}_{i,t}
        \Big)
        - \beta \,\mathbb{D}_{\mathrm{KL}}\big( \pi_{\theta} \,\|\, \pi_{\mathrm{ref}} \big)
    \Bigg\}
\Bigg] ,
\end{split}
\label{eq:GRPO-obj}
\end{equation}
where $r_{i,t} = {\pi_\theta(o_{i,t} | q, o_{i,<t})}/{\pi_{\theta_{old}}(o_{i,t} | q, o_{i,<t})}$ denotes the likelihood ratio between the current and previous policies at step $t$. The hyperparameter $\epsilon$ denoting the controls the clipping range to stabilize updates, while $\beta$ balances the main optimization objective against the KL divergence term $\mathbb{D}_\text{KL}[\pi_\theta \,||\, \pi_\text{ref}]$ which regularizes the learned policy toward a reference policy. The actual learning signal is derived from the advantage estimate, a normalized reward computed across multiple samples within the same query group:
\begin{equation}
    \hat{A}_{i,t} = \frac{r_{i,t} - \text{mean}\{r_{1,t}, \ldots, r_{N,t}\}}{\text{std}\{r_{i,t}, \ldots, r_{N,t}\}}.
\end{equation}
Where $r_{i, t}$ denotes the scalar reward assigned to the i-th generated sample

While CoT-SFT brings about the basic capability of generating long reasoning responses to the model, it does not directly guaranty superior final accuracy nor generalizability. The following RL stage tends to fail when the starting point model does not potentially master the underlying task since learning from low quality reasoning traces hardly helps further improvement. When the task is complicated, an exact match reward function will also be sparse for the model to receive meaningful training signal and hinder the converging. Unfortunately, current state of vision language models on real world 3D tasks suffers simultaneously from these complications.

In the following sections, we introduce the components used throughout our approach. First, we construct a data pipeline that produces controllable and interpretable reasoning trajectories, where text-based scene graphs serve as structured representations of spatial relations. Based on these trajectories, we apply CoT-SFT to expose the model to step-wise structured reasoning. We then describe a GRPO-based reinforcement learning stage that further refines generation quality. 

\subsection{Dataset Creation}
\label{sec:dataset}

Our training data consist of paired 3D layouts and language, organized as tuples
\((I, G_0, G^\star, x, y)\) of rendered images \(I\), an initial scene graph \(G_0\) (see
Paragraph~\ref{para:scene-graph-representation}), a target scene
graph \(G^\star\), an instruction \(x\), and a reasoning trace \(y\). The initial scene graph is obtained using a rule-based python auto-labeling framework leveraging the blender annotations.
The reasoning trace provides a
step-by-step explanation that connects the instruction and input layout to the desired final layout, and is later used for
Structured CoT-SFT.
We generate reasoning traces~$y$ using the DeepSeek-R1 model~\cite{Guo2025deepseekr1}. Given an instruction~$x$, an initial graph~$G_0$, and a target graph~$G^\star$, we prompt the model to produce a structured chain-of-thought “thinking path’’ that explicitly describes how to transform $G_0$ into $G^\star$. The generated trace maps textual constraints in the instruction—such as grouping rules, sorting criteria, relational statements, or distance specifications—to concrete object-level operations inside the scene graph.

\paragraph{Sorting task.}
The first and largest task is a synthetic 3D sorting benchmark of 10k scenes. Each instance contains a scene and a rule-heavy instruction that specifies how to group
objects and arrange them along a chosen axis. The instructions combine multiple constraints, such as (1) grouping objects by semantic attributes (e.g., shape), (2) sorting objects within each group by a geometric attribute (e.g., height), (3) placing groups in a prescribed left-to-right order, and (4) enforcing global layout constraints like a fixed total span and fixed gaps between groups and between consecutive objects. All objects are finally snapped to a common support surface, with vertical positions determined by their height and the table plane.

\paragraph{Spatial alignment task.}
The second task is a more challenging rearrangement benchmark with 1k Blender-rendered scenes used
for training. Here, the underlying target layout is a clean \(N \times M\) grid derived from an
initially well-organized scene graph, where each grid cell encodes the intended position and
orientation of an object, grouped by category. We then randomly perturb a subset of objects by
translating and rotating them away from their grid cells, yielding a corrupted initial graph
\(G_0\). The instruction asks the model to restore order by returning displaced objects to their
appropriate grid locations and orientations, while leaving correctly placed objects unchanged.

\paragraph{Room editing task.}
We further construct a room-editing benchmark based on the instruction corpus from the
InstructScene dataset~\cite{lin2024instructscene}. Although InstructScene provides rich relational
descriptions (e.g., “place the lamp next to the sofa”), such instructions generally do not specify a
unique target configuration, as many scene layouts can simultaneously satisfy the same set of
relations. Moreover, common scene-generation metrics are inherently ambiguous and fail to
sufficiently constrain the solution space. To reduce this indeterminacy, we augment each
object-insertion instruction with additional geometric constraints: for every object to be added, we
provide its distances to two or three existing reference objects. These distance constraints
significantly narrow the set of feasible placements and lead to an almost uniquely determined target
graph~$G^\star$, enabling more reliable supervision for both SFT and RL.

\subsection{3D Layout Reinforcement Learning}
\label{sec:method_rl}

We adopt a two-stage training pipeline. First, we perform CoT-SFT to provide the model with a strong cold start, ensuring it can already produce structured reasoning traces and roughly correct 3D layouts. Building upon this foundation, we then apply GRPO reinforcement learning to further refine layout accuracy, physical plausibility, and output formatting. This RL stage leverages task-specific rewards and benefits substantially from the structured behaviors acquired during CoT-SFT.

We optimize a policy to synthesize a physically plausible 3D scene graph \(G_T\) while remaining faithful to a ground-truth graph \(G^\star\). The objective is a weighted sum of three terms that jointly measure semantic alignment, internal physical feasibility, and output-format compliance:
\[
r \;=\; 
\mathrm{IoU}\!\left(G_\text{pred},\,G^\star\right)
\;+\;
\lambda_1\;\mathrm{Coll}\!\left(G_\text{pred}\right)
\;+\;
\lambda_2\;\mathrm{Fmt}\!\left(G_\text{pred}\right)
\]
Here, the weights \(\lambda_{\mathrm{Coll}}\) and \(\lambda_{\mathrm{Fmt}}\) control the relative contribution of collision avoidance and formatting fidelity, respectively.
The IoU term anchors the layout to ground-truth object geometry, the collision term suppresses physically implausible interpenetrations inside the prediction, and the format term preserves a standardized interface for downstream parsing and analysis. Jointly optimizing these signals encourages \(G_T\) to be both accurate and physically coherent while remaining reproducible and inspectable.

\paragraph{IoU reward.}
To evaluate cross-graph alignment we first establish correspondences between nodes in \(G_\text{pred}\) and \(G^\star\). For every predicted node we attempt a semantic match using textual descriptions when available; if no unique semantic match is found, we select the ground-truth node that maximizes 3D intersection over union (IoU). Let \(b_i\) and \(b_j\) be the axis-aligned 3D boxes of a matched pair. The IoU is
\(
\mathrm{IoU}(b_i,b_j) \;=\; 
{V(b_i \cap b_j)}/\left[{V(b_i) + V(b_j) - V(b_i \cap b_j)}\right],
\)
The intersection volume is computed by overlapping the one-dimensional extents on each axis. The IoU reward is the average over matched pairs,
\[
r_{\mathrm{IoU}} \;=\; \frac{1}{|G_\text{pred}|}\sum_{(i,j)\in\mathcal{M}} \mathrm{IoU}(b_i,b_j),
\]
with \(\mathcal{M}\) the set of description- or IoU-based matches.

\paragraph{Collision reward.}
To promote internal physical plausibility we penalize interpenetrations among non-container objects within \(G_T\). For every unordered pair of nodes \((p,q)\) we compute the axis-aligned intersection volume \(V_{pq}=V(b_p\cap b_q)\) using the same overlap operator. A pair is deemed colliding if \(V_{pq}>\varepsilon\) for a small tolerance \(\varepsilon>0\). Let \(\mathcal{C}=\{(p,q):V_{pq}>\varepsilon\}\) and let \(N\) be the number of non-container objects. We define a normalized collision-free score
\(
\mathrm{Coll}(G)=1-{|\mathcal{C}|}/{N},
\)
which rises as collisions vanish and reaches one when all pairs are disjoint. This term mirrors the computation used in evaluation, including the exclusion of container nodes and the use of a nonzero tolerance to avoid numerical artifacts.

\begin{table}[tb]
\centering
\resizebox{0.98\linewidth}{!}{%
\begin{tabular}{lcc|cc}
\toprule
\multirow{2}{*}{Model} &
\multicolumn{2}{c|}{Task 1: Caption-Free} &
\multicolumn{2}{c}{Task 2: Noisy-Input} \\
\cmidrule(lr){2-3}\cmidrule(lr){4-5}
& IoU & Collision Free & IoU & Collision Free\\
\midrule
Qwen3-VL-235B-Instruct & 0.408 & 0.631 & 0.428 & 0.679\\
Qwen3-VL-235B-Thinking & 0.472 & 0.744 & 0.378 & 0.701\\
\midrule
Qwen3-8B (Ours, for reference) & 0.209 & \textbf{0.984} & 0.664 & 0.939 \\
Qwen2.5-VL-7B (Ans-SFT) & 0.521 & 0.700 & 0.619 & 0.667\\
Qwen2.5-VL-7B (CoT-SFT) & 0.641 & 0.829 & 0.773 & 0.844\\
\rowcolor{gray!15}
Qwen2.5-VL-7B (Ours) & \textbf{0.787} & 0.979 & \textbf{0.822} & \textbf{0.980}
\\
\bottomrule
\end{tabular}
}
\caption{Results on caption-free and noisy-input scene-graph grounding and editing.}
\vspace{-1mm}
\label{tab:vis-dependent}
\end{table}

\paragraph{Format reward.}
To ensure outputs remain verifiable, we require a strict reasoning-and-answer pattern:
\texttt{<think> structured reasoning that includes a JSON block </think> Final Scene Graph```json\ldots```}
The format reward assigns \(\mathrm{Fmt}(G_T)=1\) when the response contains the exact tag pair \texttt{<think></think>}, at least one syntactically valid JSON block inside the \texttt{<think>} section, and a second well-formed JSON block in the final answer segment. Minor deviations such as unmatched braces or misplaced tags reduce the score continuously toward zero. This constraint stabilizes training by coupling layout predictions with auditable, structured reasoning rather than free-form prose.

\begin{table}[tbp]
\centering
\resizebox{0.98 \linewidth}{!}{%
\begin{tabular}{lcccccc}
\toprule
Model & Mean IoU $\uparrow$ & IoU@0.5 $\uparrow$ & Ctr. Dist. $\downarrow$\\
\midrule
Qwen3-235B-Instruct & 0.592 & 0.587 & 0.243\\
Qwen3-235B-Thinking & 0.445 & 0.423 & 0.320\\
Qwen3-VL-235B-Instruct & \textbf{0.786} & \textbf{0.787} & 0.177\\
Qwen3-VL-235B-Thinking & 0.623 & 0.624 & 0.189\\
Deepseek-V3 & 0.553 & 0.559 & 0.278\\
Deepseek-R1 & 0.602 & 0.607 & 0.184\\
Deepseek-R1-0528 & 0.599 & 0.606 & 0.252\\
Gemini 2.5 Flash & 0.613 & 0.611 & 0.231\\
Gemini 2.5 Pro & 0.759 & 0.772 & \textbf{0.156}\\
\midrule
Qwen2.5-VL-7B (Ans-SFT) & 0.728 & 0.725 & 0.201 \\
Qwen2.5-VL-7B (CoT-SFT) & 0.785 & 0.779 & 0.193 \\
\rowcolor{gray!15}
Qwen2.5-VL-7B (Ours) & 0.798 & 0.792 & 0.181\\
Qwen2.5-7B (Ans-SFT) & 0.733 & 0.729 & 0.201 \\
Qwen2.5-7B (CoT-SFT) & 0.764 & 0.758 & 0.189 \\
\rowcolor{gray!15}
Qwen2.5-7B (Ours) & \textbf{0.809} & \textbf{0.803} & \textbf{0.176} 
\\
Qwen3-8B (Ans-SFT) & 0.649 & 0.636 & 0.248 \\
Qwen3-8B (CoT-SFT) & 0.707 & 0.700 & 0.227\\
\rowcolor{gray!15}
Qwen3-8B (Ours) & 0.790 & 0.784 & 0.190 \\
\bottomrule
\end{tabular}
}
\caption{Quantitative results on the spatial alignment benchmark}
\label{tab:alignment}
\end{table}

\begin{table}[tb]
\centering
\resizebox{0.98\linewidth}{!}{%
\begin{tabular}{lcccc}
\toprule
Model
& Mean IoU $\uparrow$ & Ctr. Dist. $\downarrow$ & Col. Free $\uparrow$ & Edit Dist. $\downarrow$ \\
\midrule
\multicolumn{2}{l}{\textit{Large Language Models (Zero-shot)}} &&\\
Qwen3-235B-Instruct & 0.122 & 2.702 & 0.548 & 6.285 \\
Qwen3-235B-Thinking & 0.708 & 0.856 & 0.957 & 2.208 \\
Qwen3-VL-235B-Instruct & 0.145 & 2.323 & 0.421 & 8.120  \\
Qwen3-VL-235B-Thinking & 0.706 & 0.568 & 0.918 & 1.900  \\
Deepseek-V3 & 0.236 & 1.242 & 0.620 & 2.010 \\
Deepseek-R1 & 0.776 & 0.566 & 0.903 & 0.783 \\
Deepseek-R1-0528 & 0.601 & 1.163 & 0.799 & 2.021 \\[2pt]
Gemini 2.5 Flash & 0.746 & \textbf{0.458} & 0.968 & 1.840 \\
Gemini 2.5 Pro & 0.724 & 0.593 & 0.976 & 2.180 \\
LayoutGPT (Qwen3-235B) & 0.763 & 0.518 & 0.971 & 0.910 \\
LayoutGPT (Gemini-2.5) & \textbf{0.781} & 0.471 & \textbf{0.982} & \textbf{0.709}\\
\midrule
\textit{Instruct Models}&&&&\\
Qwen2.5-VL-7B (Ans-SFT) & 0.570 & 0.423 & 0.698 & 0.900 \\
Qwen2.5-VL-7B (CoT-SFT) & 0.713 & 0.330 & 0.847 & 0.853 \\
\rowcolor{gray!15}
Qwen2.5-VL-7B (Ours) & 0.798 & 0.228 & 0.961 & \textbf{0.613} 
\\

Qwen2.5-7B (Ans-SFT) & 0.622 & 0.348 & 0.745 & 1.175 \\
Qwen2.5-7B (CoT-SFT) & 0.712 & 0.345 & 0.898 & 0.992 \\
\rowcolor{gray!15}
Qwen2.5-7B (Ours) & \textbf{0.879} & \textbf{0.179} & \textbf{1.000} & 0.627 \\
\midrule

\textit{Reasoning Models}&&&\\
Cosmos-7B (Ans-SFT) & 0.501 & 0.577 & 0.683 & 1.794 \\
Cosmos-7B (CoT-SFT) & 0.641 & 0.410 & 0.757 & 1.383 \\
Cosmos-7B (Vanilla GRPO) & 0.729 & 0.297 & 0.937 & 0.811 \\
\rowcolor{gray!15}
Cosmos-7B (Ours) & 0.875 & 0.234 & 1.000 & 0.592 
\\

Qwen3-8B (Ans-SFT) & 0.552 & 0.493 & 0.700 & 1.130 \\
Qwen3-8B (CoT-SFT) & 0.635 & 0.513 & 0.854 & 1.610 \\
Qwen3 (Vanilla GRPO) & 0.850 & 0.274 & 0.954 & 0.684 \\
\rowcolor{gray!15}
Qwen3-8B (Ours) & \textbf{0.924} & \textbf{0.145} & \textbf{1.000} & \textbf{0.510} \\
\bottomrule
\end{tabular}
}
\caption{\textbf{Quantitative results on the 3D Sorting benchmark.} Models equipped with our structured scene-graph reasoning (“Ours”) consistently outperform both zero-shot LLM/VLM baselines and standard CoT-SFT fine-tuning.}
\vspace{-0.3cm}
\label{tab:sorting}
\end{table}

\section{Experiments}
\subsection{Experiment Setup}
\paragraph{Baselines.} We compare our method against two categories of state-of-the-art models. (1) \textbf{Open-Sourced Models:} We evaluate top-performing LLMs and VLMs, including the Qwen3-235B and Qwen3-VL-235B with Instruct and Thinking versions~\citep{qwen3}, and the DeepSeek series (Deepseek-V3~\citep{liu2024deepseek}, DeepSeek-R1, and DeepSeek-R1-0528~\citep{deepseekai2025deepseekr1incentivizingreasoningcapability}). (2) \textbf{Proprietary Models:} We evaluate the performance of Gemini 2.5 Pro and Gemini 2.5 Flash~\citep{comanici2025gemini} as strong, closed-source baselines.

\paragraph{Evaluation Metrics.}
We evaluate the final generated scene graph using a suite of metrics. For geometric accuracy, we compute: \textbf{Mean IoU}, the average 3D IoU between predicted and ground-truth boxes with semantic matching; \textbf{IoU@$x$}, the percentage of nodes achieving at least $x$ IoU (e.g., 0.50); and \textbf{Center Distance (Ctr. Dist.)}, the average Euclidean distance between the centroids of predicted bounding boxes and ground-truths. For physical plausibility, we report the \textbf{Collision-Free Score (Col. Free)}, which is the normalized score $\mathrm{Coll}(G)$ defined in Section~\ref{sec:method_rl}. Finally, for the sorting task, we also evaluate the \textbf{Edit Distance (Edit Dist.)}, calculated as the average Levenshtein edit distance required to transform the predicted object orders into ground-truths.

\subsection{Sorting}

\begin{figure*}[t]
    \centering
    \includegraphics[width=0.9\linewidth]{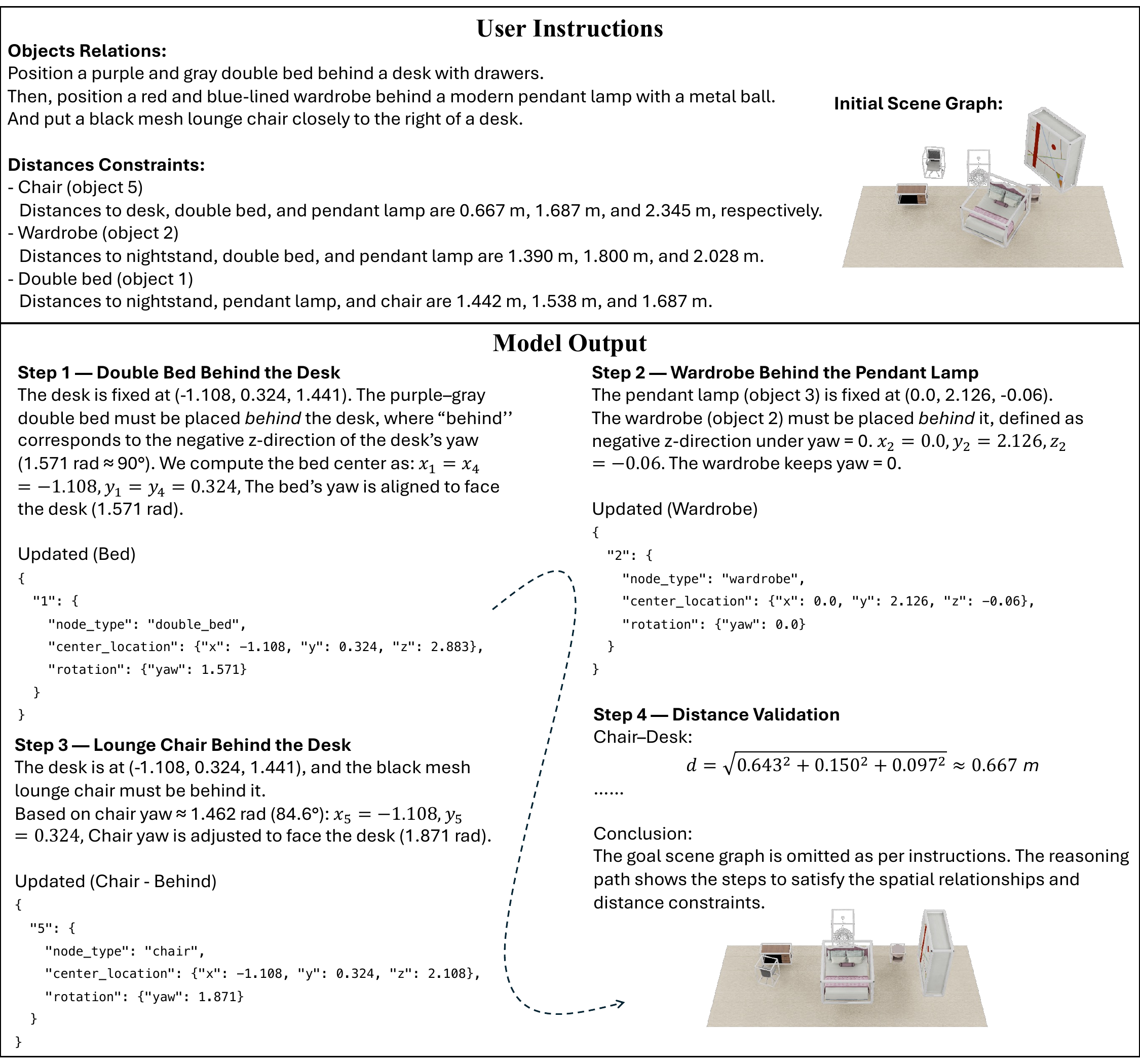}
    \caption{\textbf{Example of text-guided 3D room layout reasoning}, showing how the model interprets constraints to update object poses step by step and validate distances.}
    \vspace{-0.3cm}
    \label{fig:roomedit_qual}
\end{figure*}

In real-world or simulators, object captions may be missing, and detected bounding boxes can be inaccurate, making grounding from the scene graph ambiguous. We therefore evaluate two settings in Table~\ref{tab:vis-dependent}: (1) Caption-Free, where node captions are removed and the model must infer object semantics and relations from geometry (and visual cues when available) to recover descriptions and output the updated graph; and (2) Noisy-Input, where we perturb the input by adding 5\% jitter to object bounding boxes and enforcing partial visibility, testing robustness to imperfect localization and occlusion.
As shown in Table~\ref{tab:vis-dependent}, our method is effective with a VLM and yields strong gains under both settings.

To further reveal the capability of our structured reasoning, we additionally evaluate the sorting task under a perfect-input setting, as shown in Table~\ref{tab:sorting}. Direct SFT to generate the final scene graph consistently underperforms, as the task requires multi-step decomposition, intermediate-state perception, and recursive update—capabilities that the base model cannot acquire through SFT alone. In contrast, incorporating structured CoT-SFT enables the model to reliably construct intermediate states and reach the correct final configuration. Adding RL on top of CoT-SFT further improves performance, indicating that synthetic traces alone do not cover the harder regions of the task distribution and that RL provides additional exploration and reward-driven refinement for further improvements. To isolate the contribution of structured content, we additionally compare Vanilla GRPO training from a Qwen3-8B base model. Skipping the structured CoT-SFT stage results in a 7.4\% drop in mIoU gain, confirming that scene-graph-based trajectories offer crucial guidance.

\begin{figure*}[t]
    \centering
    \includegraphics[width=0.9\linewidth]{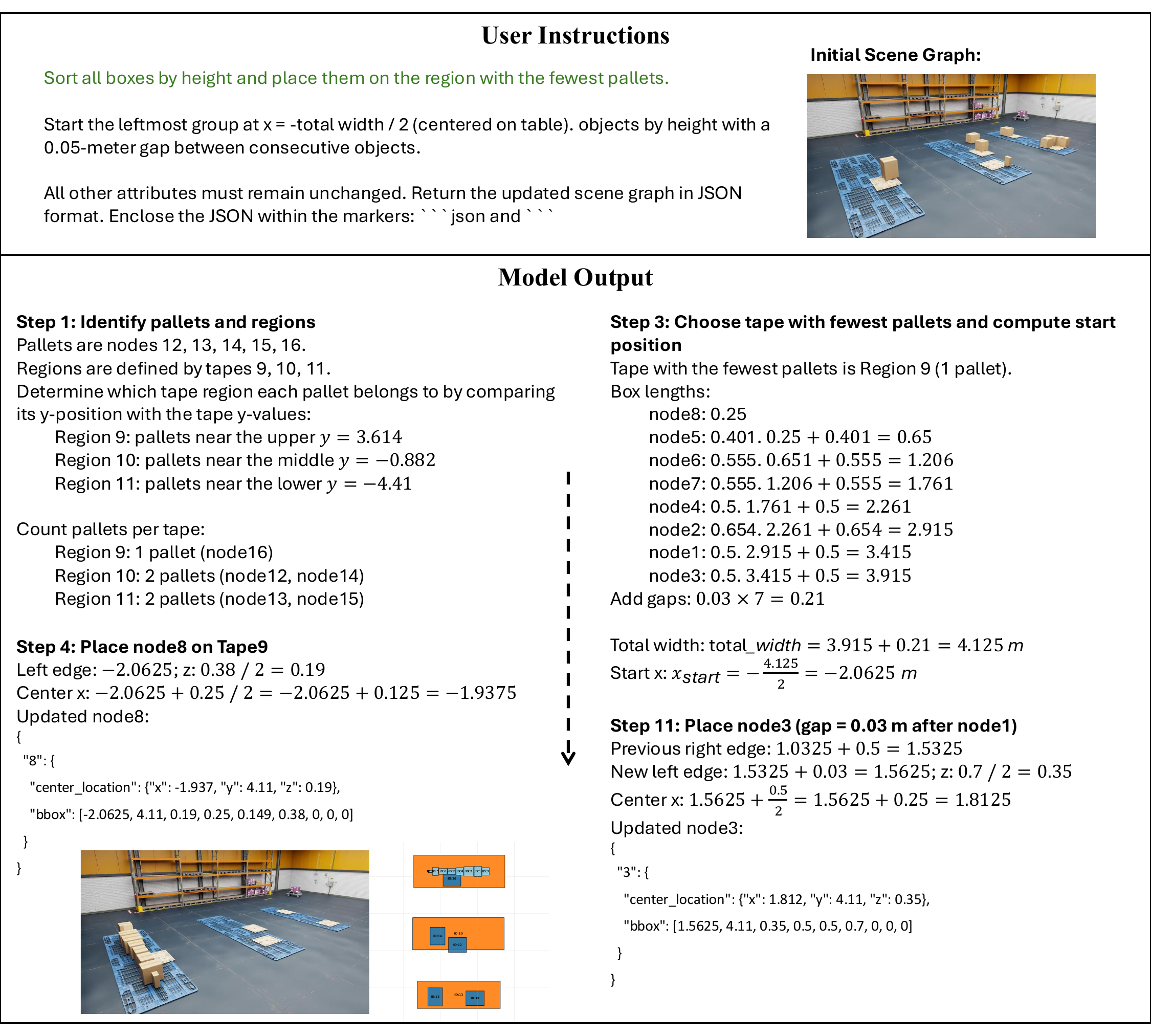}
    \caption{Out-of-domain warehouse simulation results showing our model correctly following user instructions with 3D boxes.}
    \vspace{-0.2cm}
    \label{fig:warehouse}
\end{figure*}

\textbf{Discussion.} In the perfect-input setting, the problem largely collapses to text-based structured reasoning over a clean scene graph, so adding a VLM backbone brings limited extra benefit beyond what the LLM already provides. In contrast, the core advantage of VLMs is stronger visual/geometry grounding under missing captions and imperfect perception; when captions are removed or boxes are noisy/partially visible, LLM-only models degrade much more, while VLM-based grounding remains more robust.

\subsection{Spatial Alignment}

\begin{table}[tbp]
\centering
\resizebox{0.98 \linewidth}{!}{%
\begin{tabular}{lcccccc}
\toprule
Model & Mean IoU $\uparrow$ & IoU@0.5 $\uparrow$ & Ctr. Dist. $\downarrow$\\
\midrule
Qwen3-VL-235B-Instruct & 0.502 & 0.497 & 0.719\\
Qwen3-VL-235B-Thinking & 0.496 & 0.487 & 0.684\\
Deepseek-V3 & 0.473 & 0.469 & 0.859\\
Deepseek-R1 & 0.454 & 0.449 & 0.754\\
Deepseek-R1-0528 & 0.471 & 0.463 & 0.871\\
Gemini 2.5 Flash & 0.443 & 0.441 & 0.924\\
Gemini 2.5 Pro & \textbf{0.591} & \textbf{0.587} & \textbf{0.621}\\
\midrule
Qwen2.5-7B (Ans-SFT) & 0.869 & 0.863 & 0.138 \\
Qwen2.5-7B (CoT-SFT) & 0.950 & 0.950 & 0.051 \\
\rowcolor{gray!15}
Qwen2.5-7B (Ours) & 0.974 & 0.973 & 0.033 \\
Qwen3-8B (Ans-SFT) & 0.895 & 0.897 & 0.117 \\
Qwen3-8B (CoT-SFT) & 0.972 & 0.972 & \textbf{0.026} \\
Qwen3-8B (Vanilla GRPO) & 0.768 & 0.774 & 0.336 \\
\rowcolor{gray!15}
Qwen3-8B (Ours) & \textbf{0.983} & \textbf{0.983} & 0.031 \\
\bottomrule
\end{tabular}
}
\caption{Quantitative results on the RoomEditing benchmark}
\vspace{-0.5cm}
\label{tab:room_editing}
\end{table}

The spatial alignment task poses a slightly higher difficulty than the sorting task, as it requires understanding the underlying $N \times M$ group structure, identifying inconsistencies relative to a group anchor, and locating the misaligned instance based on center-position deviations. This makes the benchmark challenging even for SOTA models, as shown in Table~\ref{tab:alignment}. 

\begin{figure*}[t]
\centering
\includegraphics[width=0.9\linewidth]{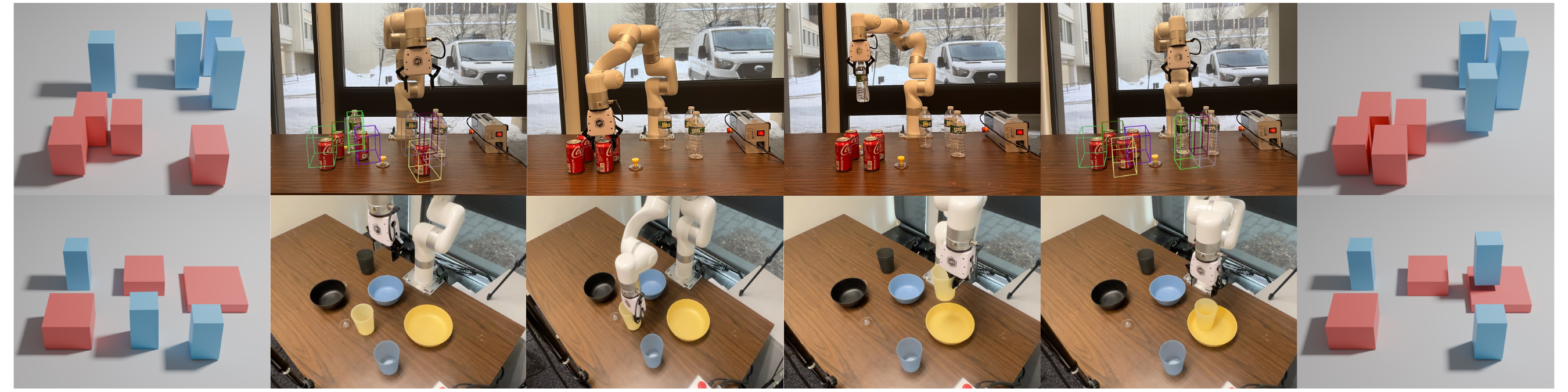}
\vspace{-0.1cm}
\caption{Real-world tabletop rearrangement and pick-and-place task.}
\vspace{-0.3cm}
\label{fig:placeholder}
\end{figure*}

From our post training experiments, answer-SFT models that directly predict the final coordinates fall noticeably short. In contrast, our structured layout reasoning can further improve the performance of smaller 7B/8B models, enabling them to match or even exceed the accuracy of these large commercial models. Strategically, our 3D-Layout-R1 learns to (1) infer the underlying grid pattern, (2) detect misaligned objects via center-location comparisons, and (3) recover the correct target position by either extending the end node or interpolating along the relevant grid line.

An additional observation is that large VLMs consistently outperform their LLM counterparts by over 10\% on this task. However, our own post-training experiments on VLM backbones do not show comparable gains, suggesting that specialized visual prompting or tailored visual representations may be necessary to fully exploit VLM capabilities. We provide further discussion and ablations in the supplementary material.

\subsection{Room Layout Editing} 
For the layout editing task, we conduct a parallel set of experiments and observe consistent trends. As shown in Table~\ref{tab:room_editing}, applying RL on top of structured CoT-SFT again produces the strongest overall performance. While the gap between direct answer SFT and methods incorporating structured reasoning persists, a notable difference emerges: for this task, the performance gap between RL and its corresponding structured CoT-SFT baseline is significantly smaller than in the sorting task. This suggests that CoT-SFT already captures most of the structured patterns required for layout editing, leaving less headroom for RL to further optimize. We attribute this to the relative simplicity of the task, each editing example modifies only a small portion of the scene, and the model primarily needs to learn how to move the specified objects to achieve a high IoU.

\subsection{Qualitative Results}
\textbf{Structured 3D sorting traces.}
Figure~\ref{fig:sorting_qual} shows an results from the synthetic sorting benchmark. Given a cluttered scene and a rule-heavy instruction, our model generates a trace that (1) groups and sorts obejcts by attributes, (2) allocates group spans along the target axis, and (3) places each object with the required gaps. Each step is accompanied by a compact scene-graph update, producing layouts that match the target.

\textbf{Room-editing reasoning.}
Figure~\ref{fig:roomedit_qual} presents a room-editing example from the InstructScene-based task.
Starting from an initial scene graph and relational instructions with metric constraints, the model
incrementally updates object poses in JSON form. The trace explicitly verifies the
Euclidean distances used in the constraints, and the resulting rendered layout is both visually
plausible and quantitatively consistent with the ground-truth scene graph.

\textbf{Warehouse.}
We additionally show out-of-domain qualitative results in a warehouse simulation. In this setting, the simulator provides perfect 3D bounding boxes, and the user specifies practical rearrangement rules. As shown in Fig.~\ref{fig:warehouse}, without any retraining on warehouse data, our model follows these unseen instructions and produces correct, physically consistent scene-graph updates.

\textbf{Real-world Robot.}
We demonstrate our method on two real-world table-top tasks: (i) a spatial alignment task, and (ii) a pick-and-place task that places the yellow cup in the yellow bowl. For both tasks, we use Qwen3-VL to extract object bounding boxes from the input image and then apply our VLM-based scene editing to synthesize the goal state. Given known grasping skills for the detected objects, a rule-based and goal-conditioned robot can execute the task. We do not explicitly model continuous dynamics or collisions inside the model; instead, our focus is high-level goal generation, which is complementary to existing motion planning and control.

\section{Conclusion}
We presented 3D-Layout-R1, a structured reasoning framework that improves the accuracy and interpretability of language-guided 3D layout editing. By injecting layout structure into multi-step reasoning and refining outputs with GRPO, our approach produces spatially consistent, physically coherent scene edits across a wide range of layout manipulation tasks. This work demonstrates the value of explicit graph-based reasoning for robust and controllable 3D understanding.
Across all benchmarks, 3D-Layout-R1 significantly reduces layout error and consistently outperforms zero-shot and CoT-based baselines, particularly in demanding spatial alignment and room-editing settings. These results highlight the framework’s strong generalization and its effectiveness in translating natural language instructions into precise 3D transformations.

\clearpage
{
    \small
    \bibliographystyle{ieeenat_fullname}
    \bibliography{main}

@article{abdelreheem2025placeit3d,
  title={PlaceIt3D: Language-Guided Object Placement in Real 3D Scenes},
  author={Abdelreheem, Ahmed and Aleotti, Filippo and Watson, Jamie and Qureshi, Zawar and Eldesokey, Abdelrahman and Wonka, Peter and Brostow, Gabriel and Vicente, Sara and Garcia-Hernando, Guillermo},
  journal={arXiv preprint arXiv:2505.05288},
  year={2025}
}

@inproceedings{el2025scanedit,
  title={ScanEdit: Hierarchically-Guided Functional 3D Scan Editing},
  author={El Amine Boudjoghra, Mohamed and Laptev, Ivan and Dai, Angela},
  booktitle={Proceedings of the IEEE/CVF International Conference on Computer Vision},
  pages={27105--27115},
  year={2025}
}

@inproceedings{gu2025blendergym,
  title={BlenderGym: Benchmarking Foundational Model Systems for Graphics Editing},
  author={Gu, Yunqi and Huang, Ian and Je, Jihyeon and Yang, Guandao and Guibas, Leonidas},
  booktitle={Proceedings of the Computer Vision and Pattern Recognition Conference},
  pages={18574--18583},
  year={2025}
}

@article{zheng2024editroom,
  title={EditRoom: LLM-parameterized Graph Diffusion for Composable 3D Room Layout Editing},
  author={Zheng, Kaizhi and Chen, Xiaotong and He, Xuehai and Gu, Jing and Li, Linjie and Yang, Zhengyuan and Lin, Kevin and Wang, Jianfeng and Wang, Lijuan and Wang, Xin Eric},
  journal={arXiv preprint arXiv:2410.12836},
  year={2024}
}

@inproceedings{huang2025fireplace,
  title={Fireplace: Geometric refinements of llm common sense reasoning for 3d object placement},
  author={Huang, Ian and Bao, Yanan and Truong, Karen and Zhou, Howard and Schmid, Cordelia and Guibas, Leonidas and Fathi, Alireza},
  booktitle={Proceedings of the Computer Vision and Pattern Recognition Conference},
  pages={13466--13476},
  year={2025}
}

@inproceedings{ling2023deductive,
 author = {Ling, Zhan and Fang, Yunhao and Li, Xuanlin and Huang, Zhiao and Lee, Mingu and Memisevic, Roland and Su, Hao},
 booktitle = {Advances in Neural Information Processing Systems},
 editor = {A. Oh and T. Naumann and A. Globerson and K. Saenko and M. Hardt and S. Levine},
 pages = {36407--36433},
 publisher = {Curran Associates, Inc.},
 title = {Deductive Verification of Chain-of-Thought Reasoning},
 url = {https://proceedings.neurips.cc/paper_files/paper/2023/file/72393bd47a35f5b3bee4c609e7bba733-Paper-Conference.pdf},
 volume = {36},
 year = {2023}
}

@article{Li25Structured,
author = {Li, Jia and Li, Ge and Li, Yongmin and Jin, Zhi},
title = {Structured Chain-of-Thought Prompting for Code Generation},
year = {2025},
issue_date = {February 2025},
publisher = {Association for Computing Machinery},
address = {New York, NY, USA},
volume = {34},
number = {2},
issn = {1049-331X},
url = {https://doi.org/10.1145/3690635},
doi = {10.1145/3690635},
journal = {ACM Trans. Softw. Eng. Methodol.},
month = jan,
articleno = {37},
numpages = {23}
}

@inproceedings{zhu2025llava,
  title={Llava-3d: A simple yet effective pathway to empowering lmms with 3d capabilities},
  author={Zhu, Chenming and Wang, Tai and Zhang, Wenwei and Pang, Jiangmiao and Liu, Xihui},
  booktitle={Proceedings of the IEEE/CVF International Conference on Computer Vision},
  pages={4295--4305},
  year={2025}
}

@article{cheng2025sr3d,
  title={3D Aware Region Prompted Vision Language Model},
  author={An-Chieh Cheng and Yang Fu and Yukang Chen and Zhijian Liu and Xiaolong Li and Subhashree Radhakrishnan and Song Han and Yao Lu and Jan Kautz and Pavlo Molchanov and Hongxu Yin and Xiaolong Wang and Sifei Liu},
  journal={arXiv preprint arXiv:2509.13317},
  year={2025},
}

@inproceedings{yang2025thinking,
  title={Thinking in space: How multimodal large language models see, remember, and recall spaces},
  author={Yang, Jihan and Yang, Shusheng and Gupta, Anjali W and Han, Rilyn and Fei-Fei, Li and Xie, Saining},
  booktitle={Proceedings of the Computer Vision and Pattern Recognition Conference},
  pages={10632--10643},
  year={2025}
}

@inproceedings{cai2025spatialbot,
  title={Spatialbot: Precise spatial understanding with vision language models},
  author={Cai, Wenxiao and Ponomarenko, Iaroslav and Yuan, Jianhao and Li, Xiaoqi and Yang, Wankou and Dong, Hao and Zhao, Bo},
  booktitle={2025 IEEE International Conference on Robotics and Automation (ICRA)},
  pages={9490--9498},
  year={2025},
  organization={IEEE}
}

@article{liu2025spatialcot,
  title={SpatialCoT: Advancing Spatial Reasoning through Coordinate Alignment and Chain-of-Thought for Embodied Task Planning},
  author={Liu, Yuecheng and Chi, Dafeng and Wu, Shiguang and Zhang, Zhanguang and Hu, Yaochen and Zhang, Lingfeng and Zhang, Yingxue and Wu, Shuang and Cao, Tongtong and Huang, Guowei and others},
  journal={arXiv preprint arXiv:2501.10074},
  year={2025}
}

@inproceedings{song2025robospatial,
  title={Robospatial: Teaching spatial understanding to 2d and 3d vision-language models for robotics},
  author={Song, Chan Hee and Blukis, Valts and Tremblay, Jonathan and Tyree, Stephen and Su, Yu and Birchfield, Stan},
  booktitle={Proceedings of the Computer Vision and Pattern Recognition Conference},
  pages={15768--15780},
  year={2025}
}

@article{zhou2025roborefer,
  title={RoboRefer: Towards Spatial Referring with Reasoning in Vision-Language Models for Robotics},
  author={Zhou, Enshen and An, Jingkun and Chi, Cheng and Han, Yi and Rong, Shanyu and Zhang, Chi and Wang, Pengwei and Wang, Zhongyuan and Huang, Tiejun and Sheng, Lu and others},
  journal={arXiv preprint arXiv:2506.04308},
  year={2025}
}

@inproceedings{chen2024spatialvlm,
  title={Spatialvlm: Endowing vision-language models with spatial reasoning capabilities},
  author={Chen, Boyuan and Xu, Zhuo and Kirmani, Sean and Ichter, Brain and Sadigh, Dorsa and Guibas, Leonidas and Xia, Fei},
  booktitle={Proceedings of the IEEE/CVF Conference on Computer Vision and Pattern Recognition},
  pages={14455--14465},
  year={2024}
}

@article{cheng2024spatialrgpt,
  title={Spatialrgpt: Grounded spatial reasoning in vision-language models},
  author={Cheng, An-Chieh and Yin, Hongxu and Fu, Yang and Guo, Qiushan and Yang, Ruihan and Kautz, Jan and Wang, Xiaolong and Liu, Sifei},
  journal={Advances in Neural Information Processing Systems},
  volume={37},
  pages={135062--135093},
  year={2024}
}

@article{Guo2025deepseekr1,
  author = {Daya Guo and Dejian Yang and Haowei Zhang and Junxiao Song and Peiyi Wang and Qihao Zhu and Runxin Xu and Ruoyu Zhang and Shirong Ma and Xiao Bi and Xiaokang Zhang and Xingkai Yu and Yu Wu and Z. F. Wu and Zhibin Gou and Zhihong Shao and Zhuoshu Li and Ziyi Gao and Aixin Liu and Bing Xue and Bingxuan Wang and Bochao Wu and Bei Feng and Chengda Lu and Chenggang Zhao and Chengqi Deng and Chong Ruan and Damai Dai and Deli Chen and Dongjie Ji and Erhang Li and Fangyun Lin and Fucong Dai and Fuli Luo and Guangbo Hao and Guanting Chen and Guowei Li and H. Zhang and Hanwei Xu and Honghui Ding and Huazuo Gao and Hui Qu and Hui Li and Jianzhong Guo and Jiashi Li and Jingchang Chen and Jingyang Yuan and Jinhao Tu and Junjie Qiu and Junlong Li and J. L. Cai and Jiaqi Ni and Jian Liang and Jin Chen and Kai Dong and Kai Hu and Kaichao You and Kaige Gao and Kang Guan and Kexin Huang and Kuai Yu and Lean Wang and Lecong Zhang and Liang Zhao and Litong Wang and Liyue Zhang and Lei Xu and Leyi Xia and Mingchuan Zhang and Minghua Zhang and Minghui Tang and Mingxu Zhou and Meng Li and Miaojun Wang and Mingming Li and Ning Tian and Panpan Huang and Peng Zhang and Qiancheng Wang and Qinyu Chen and Qiushi Du and Ruiqi Ge and Ruisong Zhang and Ruizhe Pan and Runji Wang and R. J. Chen and R. L. Jin and Ruyi Chen and Shanghao Lu and Shangyan Zhou and Shanhuang Chen and Shengfeng Ye and Shiyu Wang and Shuiping Yu and Shunfeng Zhou and Shuting Pan and S. S. Li and Shuang Zhou and Shaoqing Wu and Tao Yun and Tian Pei and Tianyu Sun and T. Wang and Wangding Zeng and Wen Liu and Wenfeng Liang and Wenjun Gao and Wenqin Yu and Wentao Zhang and W. L. Xiao and Wei An and Xiaodong Liu and Xiaohan Wang and Xiaokang Chen and Xiaotao Nie and Xin Cheng and Xin Liu and Xin Xie and Xingchao Liu and Xinyu Yang and Xinyuan Li and Xuecheng Su and Xuheng Lin and X. Q. Li and Xiangyue Jin and Xiaojin Shen and Xiaosha Chen and Xiaowen Sun and Xiaoxiang Wang and Xinnan Song and Xinyi Zhou and Xianzu Wang and Xinxia Shan and Y. K. Li and Y. Q. Wang and Y. X. Wei and Yang Zhang and Yanhong Xu and Yao Li and Yao Zhao and Yaofeng Sun and Yaohui Wang and Yi Yu and Yichao Zhang and Yifan Shi and Yiliang Xiong and Ying He and Yishi Piao and Yisong Wang and Yixuan Tan and Yiyang Ma and Yiyuan Liu and Yongqiang Guo and Yuan Ou and Yuduan Wang and Yue Gong and Yuheng Zou and Yujia He and Yunfan Xiong and Yuxiang Luo and Yuxiang You and Yuxuan Liu and Yuyang Zhou and Y. X. Zhu and Yanping Huang and Yaohui Li and Yi Zheng and Yuchen Zhu and Yunxian Ma and Ying Tang and Yukun Zha and Yuting Yan and Z. Z. Ren and Zehui Ren and Zhangli Sha and Zhe Fu and Zhean Xu and Zhenda Xie and Zhengyan Zhang and Zhewen Hao and Zhicheng Ma and Zhigang Yan and Zhiyu Wu and Zihui Gu and Zijia Zhu and Zijun Liu and Zilin Li and Ziwei Xie and Ziyang Song and Zizheng Pan and Zhen Huang and Zhipeng Xu and Zhongyu Zhang and Zhen Zhang},
  title = {DeepSeek-R1 incentivizes reasoning in LLMs through reinforcement learning},
  journal = {Nature},
  year = {2025},
  volume = {645},
  number = {8081},
  pages = {633--638},
  doi = {10.1038/s41586-025-09422-z},
  url = {https://doi.org/10.1038/s41586-025-09422-z},
}

@article{qwen3,
    title={Qwen3 Technical Report}, 
    author={An Yang and Anfeng Li and Baosong Yang and Beichen Zhang and Binyuan Hui and Bo Zheng and Bowen Yu and Chang Gao and Chengen Huang and Chenxu Lv and Chujie Zheng and Dayiheng Liu and Fan Zhou and Fei Huang and Feng Hu and Hao Ge and Haoran Wei and Huan Lin and Jialong Tang and Jian Yang and Jianhong Tu and Jianwei Zhang and Jianxin Yang and Jiaxi Yang and Jing Zhou and Jingren Zhou and Junyang Lin and Kai Dang and Keqin Bao and Kexin Yang and Le Yu and Lianghao Deng and Mei Li and Mingfeng Xue and Mingze Li and Pei Zhang and Peng Wang and Qin Zhu and Rui Men and Ruize Gao and Shixuan Liu and Shuang Luo and Tianhao Li and Tianyi Tang and Wenbiao Yin and Xingzhang Ren and Xinyu Wang and Xinyu Zhang and Xuancheng Ren and Yang Fan and Yang Su and Yichang Zhang and Yinger Zhang and Yu Wan and Yuqiong Liu and Zekun Wang and Zeyu Cui and Zhenru Zhang and Zhipeng Zhou and Zihan Qiu},
    journal = {arXiv preprint arXiv:2505.09388},
    year={2025}
}

@article{huang2025frag,
  title={FRAG: Frame Selection Augmented Generation for Long Video and Long Document Understanding},
  author={De-An Huang and Subhashree Radhakrishnan and Zhiding Yu and Jan Kautz},
  journal={arXiv preprint arXiv:2504.17447},
  year={2025}
}

@article{Anurag2025Temporal,
  author       = {Anurag Arnab and
                  Ahmet Iscen and
                  Mathilde Caron and
                  Alireza Fathi and
                  Cordelia Schmid},
  title        = {Temporal Chain of Thought: Long-Video Understanding by Thinking in
                  Frames},
  journal      = {CoRR},
  volume       = {abs/2507.02001},
  year         = {2025},
  url          = {https://doi.org/10.48550/arXiv.2507.02001},
  doi          = {10.48550/ARXIV.2507.02001},
  eprinttype    = {arXiv},
  eprint       = {2507.02001},
  bibsource    = {dblp computer science bibliography, https://dblp.org}
}

@misc{guo2025structuredoutputsenablegeneralpurpose,
      title={Structured Outputs Enable General-Purpose LLMs to be Medical Experts}, 
      author={Guangfu Guo and Kai Zhang and Bryan Hoo and Yujun Cai and Xiaoqian Lu and Nanyun Peng and Yiwei Wang},
      year={2025},
      eprint={2503.03194},
      archivePrefix={arXiv},
      primaryClass={cs.CL},
      url={https://arxiv.org/abs/2503.03194}, 
}

@inproceedings{Fei2024Video,
author = {Fei, Hao and Wu, Shengqiong and Ji, Wei and Zhang, Hanwang and Zhang, Meishan and Lee, Mong Li and Hsu, Wynne},
title = {Video-of-thought: step-by-step video reasoning from perception to cognition},
year = {2024},
publisher = {JMLR.org},
booktitle = {Proceedings of the 41st International Conference on Machine Learning},
articleno = {526},
numpages = {17},
location = {Vienna, Austria},
series = {ICML'24}
}

@misc{deepseek-math,
  author = {Zhihong Shao and Peiyi Wang and Qihao Zhu and Runxin Xu and Junxiao Song and Mingchuan Zhang and Y.K. Li and Y. Wu and Daya Guo},
  title = {DeepSeekMath: Pushing the Limits of Mathematical Reasoning in Open Language Models},
  journal = {CoRR},
  volume = {abs/2402.03300},
  year = {2024},
  url = {https://arxiv.org/abs/2402.03300},
}

@inproceedings{li2025RaLU,
  author       = {Cheryl Li and Tianyuan Xu and Ryman Guo},
  title        = {Reasoning-as-Logic-Units: Scaling Test-Time Reasoning in Large Language Models Through Logic Unit Alignment},
  booktitle    = {International Conference on Machine Learning, {ICML} 2025, 13-19 July 2025, Vancouver, Canada},
  series       = {Proceedings of Machine Learning Research},
  publisher    = {{PMLR}},
  year         = {2025},
  url          = {[https://proceedings.mlr.press/v202/gao23f.html](https://www.arxiv.org/abs/2502.07803)},
}

@article{yu2025dapo,
  title={Dapo: An open-source llm reinforcement learning system at scale},
  author={Yu, Qiying and Zhang, Zheng and Zhu, Ruofei and Yuan, Yufeng and Zuo, Xiaochen and Yue, Yu and Dai, Weinan and Fan, Tiantian and Liu, Gaohong and Liu, Lingjun and others},
  journal={arXiv preprint arXiv:2503.14476},
  year={2025}
}

@article{gspo,
  title={Group Sequence Policy Optimization}, 
  author={
    Chujie Zheng and Shixuan Liu and Mingze Li and Xiong-Hui Chen and Bowen Yu and 
    Chang Gao and Kai Dang and Yuqiong Liu and Rui Men and An Yang and Jingren Zhou and 
    Junyang Lin 
  },
  journal={arXiv preprint arXiv:2507.18071},
  year={2025}
}

@article{huang2025survey,
  title={A survey on hallucination in large language models: Principles, taxonomy, challenges, and open questions},
  author={Huang, Lei and Yu, Weijiang and Ma, Weitao and Zhong, Weihong and Feng, Zhangyin and Wang, Haotian and Chen, Qianglong and Peng, Weihua and Feng, Xiaocheng and Qin, Bing and others},
  journal={ACM Transactions on Information Systems},
  volume={43},
  number={2},
  pages={1--55},
  year={2025},
  publisher={ACM New York, NY}
}

@inproceedings{ma2025spatialllm,
  title={Spatialllm: A compound 3d-informed design towards spatially-intelligent large multimodal models},
  author={Ma, Wufei and Ye, Luoxin and de Melo, Celso M and Yuille, Alan and Chen, Jieneng},
  booktitle={Proceedings of the Computer Vision and Pattern Recognition Conference},
  pages={17249--17260},
  year={2025}
}

@article{ma2025spatialreasoner,
  title={Spatialreasoner: Towards explicit and generalizable 3d spatial reasoning},
  author={Ma, Wufei and Chou, Yu-Cheng and Liu, Qihao and Wang, Xingrui and de Melo, Celso and Xie, Jianwen and Yuille, Alan},
  journal={arXiv preprint arXiv:2504.20024},
  year={2025}
}

@article{liu2024deepseek,
  title={Deepseek-v3 technical report},
  author={Liu, Aixin and Feng, Bei and Xue, Bing and Wang, Bingxuan and Wu, Bochao and Lu, Chengda and Zhao, Chenggang and Deng, Chengqi and Zhang, Chenyu and Ruan, Chong and others},
  journal={arXiv preprint arXiv:2412.19437},
  year={2024}
}

@misc{deepseekai2025deepseekr1incentivizingreasoningcapability,
      title={DeepSeek-R1: Incentivizing Reasoning Capability in LLMs via Reinforcement Learning}, 
      author={DeepSeek-AI},
      year={2025},
      eprint={2501.12948},
      archivePrefix={arXiv},
      primaryClass={cs.CL},
      url={https://arxiv.org/abs/2501.12948}, 
}

@article{comanici2025gemini,
  title={Gemini 2.5: Pushing the frontier with advanced reasoning, multimodality, long context, and next generation agentic capabilities},
  author={Comanici, Gheorghe and Bieber, Eric and Schaekermann, Mike and Pasupat, Ice and Sachdeva, Noveen and Dhillon, Inderjit and Blistein, Marcel and Ram, Ori and Zhang, Dan and Rosen, Evan and others},
  journal={arXiv preprint arXiv:2507.06261},
  year={2025}
}

@inproceedings{yang2024holodeck,
  title={Holodeck: Language guided generation of 3d embodied ai environments},
  author={Yang, Yue and Sun, Fan-Yun and Weihs, Luca and VanderBilt, Eli and Herrasti, Alvaro and Han, Winson and Wu, Jiajun and Haber, Nick and Krishna, Ranjay and Liu, Lingjie and others},
  booktitle={Proceedings of the IEEE/CVF Conference on Computer Vision and Pattern Recognition},
  pages={16227--16237},
  year={2024}
}

@article{sarch2025grounded,
  title={Grounded Reinforcement Learning for Visual Reasoning},
  author={Sarch, Gabriel and Saha, Snigdha and Khandelwal, Naitik and Jain, Ayush and Tarr, Michael J and Kumar, Aviral and Fragkiadaki, Katerina},
  journal={arXiv preprint arXiv:2505.23678},
  year={2025}
}

@inproceedings{sun2025layoutvlm,
  title={Layoutvlm: Differentiable optimization of 3d layout via vision-language models},
  author={Sun, Fan-Yun and Liu, Weiyu and Gu, Siyi and Lim, Dylan and Bhat, Goutam and Tombari, Federico and Li, Manling and Haber, Nick and Wu, Jiajun},
  booktitle={Proceedings of the Computer Vision and Pattern Recognition Conference},
  pages={29469--29478},
  year={2025}
}

@article{spatialreasoner,
  title={SpatialReasoner: Towards Explicit and Generalizable 3D Spatial Reasoning},
  author={Ma, Wufei and Chou, Yu-Cheng and Liu, Qihao and Wang, Xingrui and de Melo, Celso and Xie, Jianwen and Yuille, Alan},
  journal={arXiv preprint arXiv:2504.20024},
  year={2025},
  doi={10.48550/arXiv.2504.20024}
}

@inproceedings{yang2025optiscene,
  title={OptiScene: LLM-driven Indoor Scene Layout Generation via Scaled Human-aligned Data Synthesis and Multi-Stage Preference Optimization},
  author={Yang, Yixuan and Luo, Zhen and Ding, Tongsheng and Lu, Junru and Gao, Mingqi and Yang, Jinyu and Sanchez, Victor and Zheng, Feng},
  booktitle={The Thirty-ninth Annual Conference on Neural Information Processing Systems},
  year={2025}
}

@article{zhou2024layout,
  title={Layout-your-3d: Controllable and precise 3d generation with 2d blueprint},
  author={Zhou, Junwei and Li, Xueting and Qi, Lu and Yang, Ming-Hsuan},
  journal={arXiv preprint arXiv:2410.15391},
  year={2024}
}

@article{lin2024instructscene,
  title={Instructscene: Instruction-driven 3d indoor scene synthesis with semantic graph prior},
  author={Lin, Chenguo and Mu, Yadong},
  journal={arXiv preprint arXiv:2402.04717},
  year={2024}
}

@article{huang20253d,
  title={3d-r1: Enhancing reasoning in 3d vlms for unified scene understanding},
  author={Huang, Ting and Zhang, Zeyu and Tang, Hao},
  journal={arXiv preprint arXiv:2507.23478},
  year={2025}
}

@article{feng2023layoutgpt,
  title={Layoutgpt: Compositional visual planning and generation with large language models},
  author={Feng, Weixi and Zhu, Wanrong and Fu, Tsu-jui and Jampani, Varun and Akula, Arjun and He, Xuehai and Basu, Sugato and Wang, Xin Eric and Wang, William Yang},
  journal={Advances in Neural Information Processing Systems},
  volume={36},
  pages={18225--18250},
  year={2023}
}

@article{ran2025direct,
  title={Direct Numerical Layout Generation for 3D Indoor Scene Synthesis via Spatial Reasoning},
  author={Ran, Xingjian and Li, Yixuan and Xu, Linning and Yu, Mulin and Dai, Bo},
  journal={arXiv preprint arXiv:2506.05341},
  year={2025}
}

@inproceedings{daxberger2025mm,
  title={Mm-spatial: Exploring 3d spatial understanding in multimodal llms},
  author={Daxberger, Erik and Wenzel, Nina and Griffiths, David and Gang, Haiming and Lazarow, Justin and Kohavi, Gefen and Kang, Kai and Eichner, Marcin and Yang, Yinfei and Dehghan, Afshin and others},
  booktitle={Proceedings of the IEEE/CVF International Conference on Computer Vision},
  pages={7395--7408},
  year={2025}
}

@inproceedings{zheng2025video,
  title={Video-3d llm: Learning position-aware video representation for 3d scene understanding},
  author={Zheng, Duo and Huang, Shijia and Wang, Liwei},
  booktitle={Proceedings of the Computer Vision and Pattern Recognition Conference},
  pages={8995--9006},
  year={2025}
}
}
\clearpage
\setcounter{page}{1}
\setcounter{section}{0}

\maketitlesupplementary

\section{Implementation Details}

\paragraph{Base Models.}
We build on two instruction-tuned models: a vision--language model (VLM), Qwen2.5-VL-7B-Instruct, and a text-only Qwen3-Instruct model.

\paragraph{CoT Supervised Fine-tuning.}
In the first stage, we perform chain-of-thought (CoT) supervised fine-tuning on our reasoning-augmented dataset.
For the VLM (Qwen2.5-VL-7B-Instruct), we freeze the ViT-based vision backbone and only update the multimodal fusion and language components; for Qwen3 we train the text backbone with the same optimization settings.
We fine-tune for 5 epochs with a global batch size of 16 (per-device batch size 1 on 16 A100 GPUs with a gradient accumulation step of 2).
We use AdamW with a learning rate of $2\times10^{-7}$, no weight decay, and a constant-with-warmup schedule (20 warmup steps).
Training is performed in bfloat16 precision with gradient checkpointing and ZeRO-3-style sharding for memory efficiency.
We set the maximum sequence length to 10{,}240 tokens and dynamically resize images such that the number of input pixels lies between 784 and 50{,}176, following the Qwen2.5-VL vision tokenizer.

\paragraph{Reinforcement Learning Fine-tuning.}
The second stage applies reinforcement learning.
We initialize the policy from the CoT-SFT checkpoint and optimize it using a GRPO variant of PPO implemented in the \emph{verl} framework.
We train on 8 GPUs with a global batch size of 32, a learning rate of $1\times10^{-6}$, and a total of 15 epochs.
During rollouts, we allow up to 8{,}192 tokens for the prompt and up to 16{,}384 tokens for the generated response, enabling long-context reasoning.
We adopt a KL-regularized objective with a small coefficient ($\lambda_{\mathrm{KL}} = 0.01$) to keep the policy close to the reference model while still allowing behavior improvement.
Distributed training uses FSDP2 with sequence parallelism of size 2 and gradient checkpointing to reduce memory usage, and rollouts are generated with a vLLM-style engine with multiple samples per prompt.

\begin{figure}[tb]
  \centering
  \begin{tcolorbox}[width=0.99\linewidth,
                    colback=gray!5,
                    colframe=black!40,
                    arc=2mm,
                    boxrule=0.4pt]
    \scriptsize\setstretch{1.2}
    \textbf{System role:} You are a spatially intelligent, embodied AI brain specialized in spatial and interactive understanding, tasked with generating reasoning path in response to the user's queries. The user provides commands or questions related to spatial intelligence, often with a initial scene graph and a goal scene graph. Your job is to analyze the given instruction and provide a reasoning trace, which is a list of actions that transform the initial scene graph to the goal scene graph.
 \\[0.25em]
    \textbf{Inputs:}
    You will receive the following:
    \\
    1. Question: A user's question about spatial relationships, object properties, or arrangements within the scene.
    \\
    2. Initial Scene Graph: (JSON format) Structured data providing detailed geometric and semantic information about objects in the scene
    \\
    3. Goal Scene Graph: (JSON format) Structured data providing detailed geometric and semantic information about objects in the scene, organized same as the initial scene graph.
    \\
    4. (optional) distance constraints for the goal layout. \\[0.25em]
    \textbf{Guidelines:}
    \\
    1. Identify all significant objects in the image and the question, find relevant objects that are needed to solve the question.
    \\
    2. Prefer to select more relevant objects rather than miss any that could be necessary. Keep objects that may be relevant to the question.
    \\
    3. You can NOT see the goal scene graph when reasoning, you need to reason about the goal scene graph based on the initial scene graph and the question.
 \\[0.25em]
    \textbf{Outputs:}
    \\
    1. Match the relevant objects with the scene graph data, and output these objects' names and their corresponding spatial information in the scene graph format.
    \\
    2. In your reasoning path, use updated scene graph with same json format for each step if needed.
    \\
    3. Output the whole reasoning path, each step should be a valid json sub-graph or a modified json object with a detailed thinking process or a math expression in text.
    \\
    4. Omit the final step of the reasoning path, which is the goal scene graph.
    \\
    5. Format: Step i: detailed thinking process or math expression in text + json sub-graph.
    \\
    6. Remember to think about rearrengement steps. E.g., how to place the object in the position without collision.
  \end{tcolorbox}
  \caption{Template of the prompts used to generate reasoning traces over scene graphs for all spatial tasks.}
  \label{fig:reasoning-prompt}
\end{figure}

\paragraph{Inference.}
At evaluation inference time, we use a deterministic decoding strategy to better measure reasoning ability.
Specifically, we use greedy decoding with temperature set to 0.0, top-p fixed at 1.0, top-k disabled (set to $-1$), minimum probability threshold set to 0.0, and best-of set to 1.
Unless otherwise noted, we use the same decoding configuration for both Qwen2.5-VL and Qwen3 models.

\section{Reasoning Prompt Design for Data Creation}

To synthesize chain-of-thought traces over scene graphs, we design three closely related prompts for (i) generic spatial reasoning, (ii) distance-constrained InstructScene reasoning, and (iii) grid-based arrangement. All prompts share a common structure and only differ in a small set of task-specific instructions.

\begin{figure*}[t]
    \centering
    \includegraphics[width=0.98\linewidth]{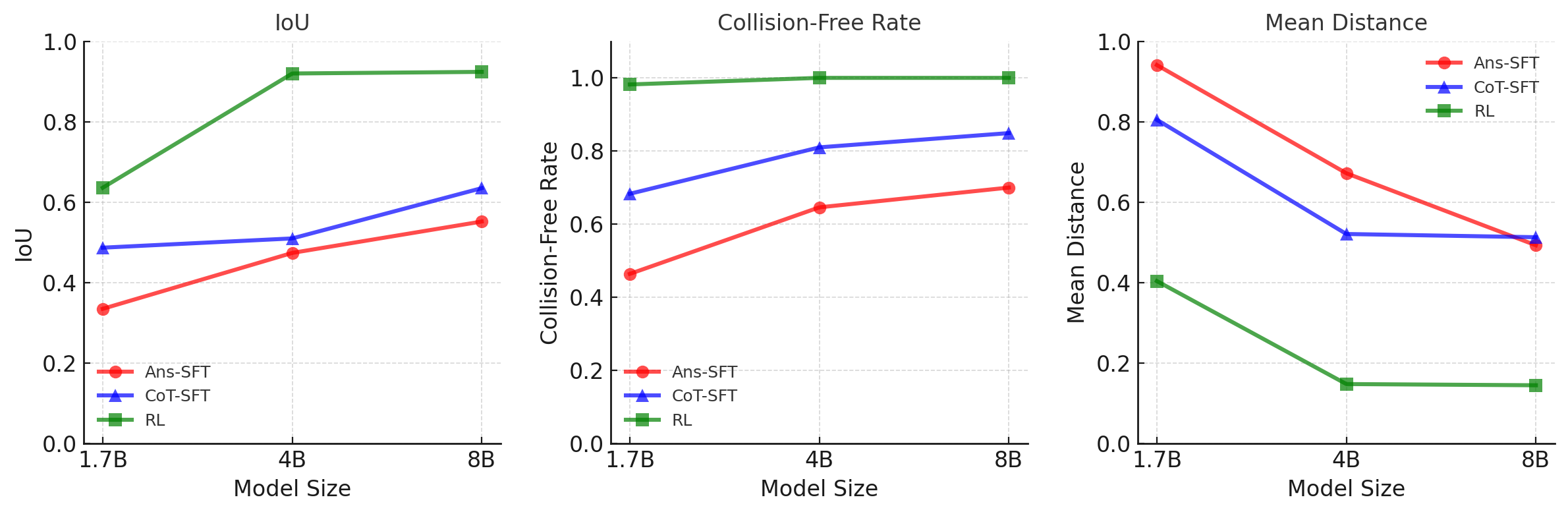}
    \caption{\textbf{Scaling behavior on the Sorting benchmark.} Reasoning-based training improves performance consistently across all model scales (1.7B--8B).}
    \label{fig:sorting_curves}
\end{figure*}

All three tasks frame the model as an embodied spatial agent that must transform an initial scene graph into a hidden goal graph. Each prompt gives: (1) a natural-language spatial question, (2) the initial JSON scene graph, and (3) the goal-format schema (but the goal contents are not directly accessible). The model must infer the goal from the question and initial state, then produce a multi-step trace where each step includes a brief explanation plus an updated/partial JSON subgraph; the final goal graph is never shown. Shared rules: include all relevant objects, ensure collision-free placements, and keep coordinates consistent. Differences: the generic prompt uses only question + graphs; InstructScene adds per-object distance constraints and requires explicit distance calculations in the trace; the grid prompt fixes the task to arranging objects into an \(N \times M\) based grid, outputting updates only for mispositioned items.

\section{More Experiments}

The scaling trends in Figure \ref{fig:sorting_curves} show that RL consistently outperforms the other training paradigms across model sizes and exhibits more favorable scaling behavior overall.

To further understand the contribution of each reward component, we conduct an ablation study on the reward design. The results in Table \ref{tab:more-exp} show that the Format and IoU rewards are crucial for strong performance, while the CF reward provides a comparatively modest but still consistent gain.

\begin{table}[H]
\centering
\resizebox{0.98 \linewidth}{!}{%
\begin{tabular}{ccc|ccc}
\toprule
Format & IoU & \ \ CF\ \ & \ \ \ \ IoU\ \ \ \  & Collision Free & Center Distance \\
\midrule
\xmark & \cmark & \cmark & 0.819 & 1.000 & 0.274 \\
\cmark & \xmark & \cmark & 0.415 & 1.000 & 0.998 \\
\cmark & \cmark & \xmark & 0.919 & 0.981 & 0.153 \\
\cmark & \cmark & \cmark & \textbf{0.924} & 1.000 & \textbf{0.145} \\
\bottomrule
\end{tabular}
}
\caption{\textbf{Ablation on reward components.} Format and IoU rewards are essential, while CF brings an additional but smaller improvement.}
\label{tab:more-exp}
\end{table}

\section{More Qualitative Results}

\begin{figure*}[tb]
    \centering
    \includegraphics[width=0.98\linewidth]{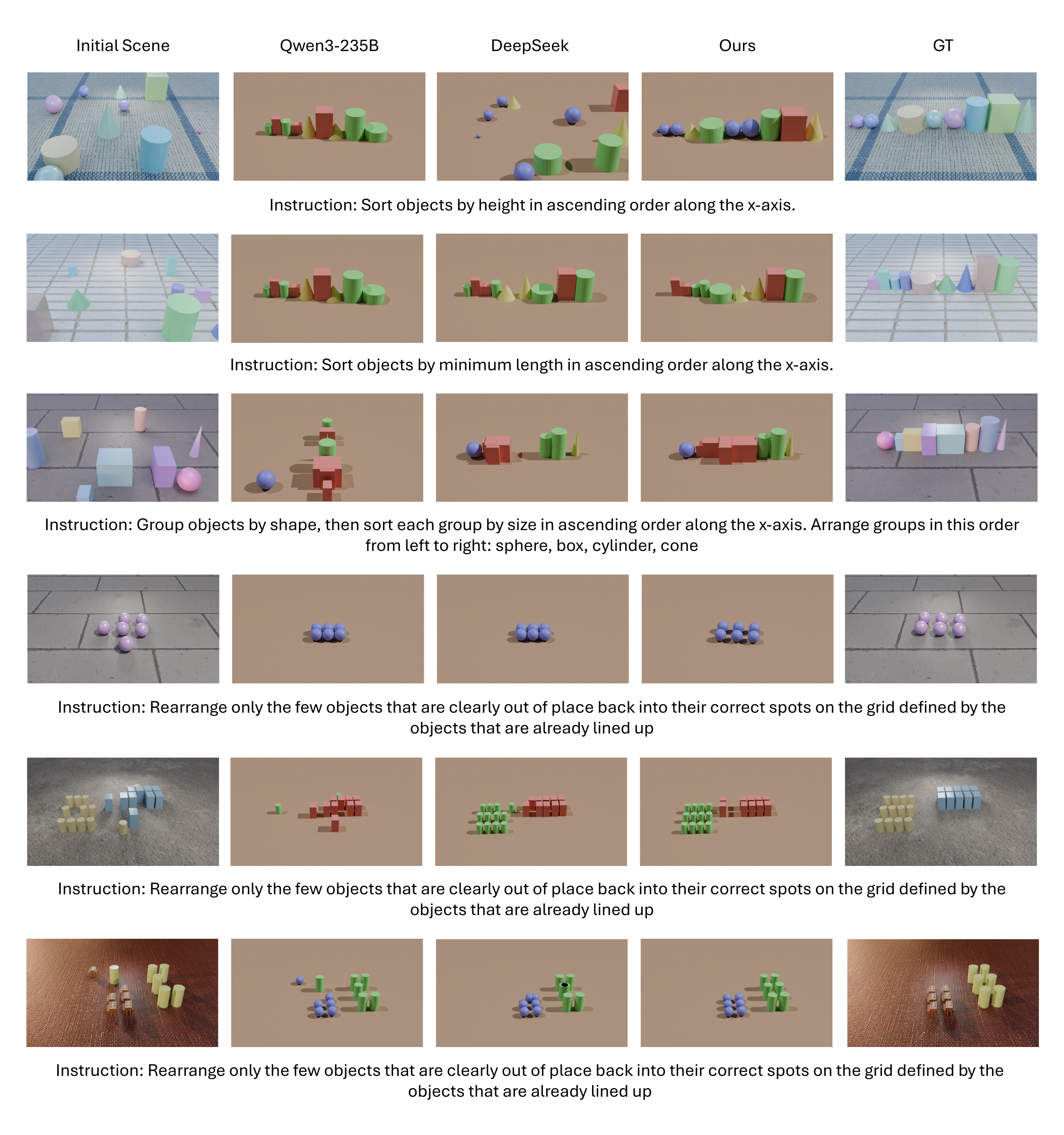}
    \caption{Qualitative comparison of sorting and spatial alignment tasks: Qwen3-235B and DeepSeek often fail to follow the instructions and produce collisions/overlaps, while our method matches the ground truth.}
    \label{fig:indomain-qual}
\end{figure*}

Figure~\ref{fig:indomain-qual} provides a qualitative comparison on sorting and alignment scenes. In each row, the initial cluttered scene is followed by the outputs of Qwen3-235B and DeepSeek baselines, our method, and the ground-truth target. While the target layouts require multi-constraint reasoning (e.g., grouping by shape, sorting by geometric attributes, preserving gaps, and snapping to a latent grid), Qwen3 and DeepSeek frequently misinterpret or only partially follow these instructions. Typical failure modes include incorrect group ordering, violating the specified ascending criteria, and drifting objects off the intended axis/grid. More importantly, their generations are often physically inconsistent: objects overlap or collide, and spacing constraints are ignored, producing implausible interpenetrations.

\end{document}